\documentclass[conference]{IEEEtran}
\IEEEoverridecommandlockouts
\usepackage{cite}
\usepackage{amsmath,amssymb,amsfonts}
\usepackage{algorithmic}
\usepackage{url} 
\usepackage{graphicx}
\usepackage{textcomp}
\usepackage{xcolor}
\def\BibTeX{{\rm B\kern-.05em{\sc i\kern-.025em b}\kern-.08em
    T\kern-.1667em\lower.7ex\hbox{E}\kern-.125emX}}
\begin{document}

\title{Knowledge-Infused LLM-Powered Conversational Health Agent: A Case Study for Diabetes Patients
% \thanks{*Authors equally contributed.}
 }

\author{\IEEEauthorblockN{Mahyar Abbasian$^1$, Zhongqi Yang$^1$, Elahe Khatibi$^1$, Pengfei Zhang$^1$, Nitish Nagesh$^1$,  \\Iman Azimi$^1$,  Ramesh Jain$^1$, and Amir M. Rahmani$^{1,2}$\\
\textit{$^1$Department of Computer Science, University of California, Irvine}\\
\textit{$^2$School of Nursing, University of California, Irvine}\\
}
\IEEEauthorblockA{
\{abbasiam, zhongqy4, ekhatibi, pengfz5, nnagesh1, azimii, rcjain, a.rahmani\}@uci.edu}}

\maketitle

\begin{abstract}
Effective diabetes management is crucial for maintaining health in diabetic patients. Large Language Models (LLMs) have opened new avenues for diabetes management, facilitating their efficacy. However, current LLM-based approaches are limited by their dependence on general sources and lack of integration with domain-specific knowledge, leading to inaccurate responses. In this paper, we propose a knowledge-infused LLM-powered conversational health agent (CHA) for diabetic patients. We customize and leverage the open-source openCHA framework, enhancing our CHA with external knowledge and analytical capabilities. This integration involves two key components: 1) incorporating the American Diabetes Association dietary guidelines and the Nutritionix information and 2) deploying analytical tools that enable nutritional intake calculation and comparison with the guidelines. We compare the proposed CHA with GPT4. Our evaluation includes 100 diabetes-related questions on daily meal choices and assessing the potential risks associated with the suggested diet. Our findings show that the proposed agent demonstrates superior performance in generating responses to manage essential nutrients.

\end{abstract}


\section{Introduction}

Effective diabetes management plays a pivotal role in maintaining optimal health in individuals diagnosed with diabetes. Given the widespread prevalence of diabetes and its significant impact on both global healthcare infrastructures and personal health outcomes, the availability and utilization of robust management strategies are essential~\cite{poulsen2010diabetes, mertz2018automated, klein2004self}. Such services require holistic approaches integrating a healthy lifestyle, nutritious diet, and physical activity assessment~\cite{fico2014integration}. Particularly, a critical aspect is dietary regulation, which plays a direct role in controlling blood glucose levels and influencing the progression of the disease~\cite{Shalahuddin2022Blood,Russell2013Nutritional}. The recent surge in advanced technologies, particularly in Large Language Models (LLMs), has substantially enhanced the accessibility and efficacy of diabetes management as transformative educational and assistive tools, enabling patients to acquire comprehensive knowledge and bridge the gap in diabetes self-management~\cite{sng2023potential,yang2023exploring,sun2023ai}.

Recent studies have investigated the implementation of LLMs on diabetes management~\cite{sng2023potential,yang2023exploring,sun2023ai}. For example,~\cite{sng2023potential} evaluated the use of ChatGPT for Diabetes Self-Management and Education. They leveraged ChatGPT to answer a range of diabetes self-management questions, covering four areas: diet/exercise, glucose level education, insulin storage, and administration. ChatGPT responded to the inquiries, showcasing a structured response style and offering guidance that is understandable to laymen. However, in some specific contexts, such as devising dietary plans, it still requires extra prompts to comprehensively produce guidelines for tasks like insulin administration. Additionally, Yang et al.~\cite{yang2023exploring} introduced ChatGLM to provide diabetes treatment strategies. The proposed model was fine-tuned through P-tuning~\cite{liu2022p} and LoRA~\cite{hu2021lora} techniques with Electronic Health Record (EHR) of patients with diabetes. ChatGLM showed proficiency in generating treatment recommendations for most cases. However, its dependency on the specific data used for fine-tuning could result in misleading outputs, potentially harmful in the absence of certain essential external diabetes knowledge.

The lack of integration with verified domain-specific knowledge underscores a significant gap in existing approaches, compromising the accuracy and reliability of their outputs. Unlike systems anchored in specific knowledge bases, existing LLM-based approaches fall into limitations in their reliance on general information sources rather than specialized for diabetes. For instance, ChatGPT's foundation on a general database -- rather than medically specialized and validated ones -- might account for its shortcomings in grasping medical nuances. Furthermore, due to the model's inability to validate the reliability of its information, they may cause ``hallucination," generating inaccurate responses presented persuasively and fluently. This problem could easily mislead individuals lacking prior knowledge on the topic~\cite{sng2023potential}.

\begin{figure}[!b]
    \centering
    \includegraphics[width=0.4\textwidth]{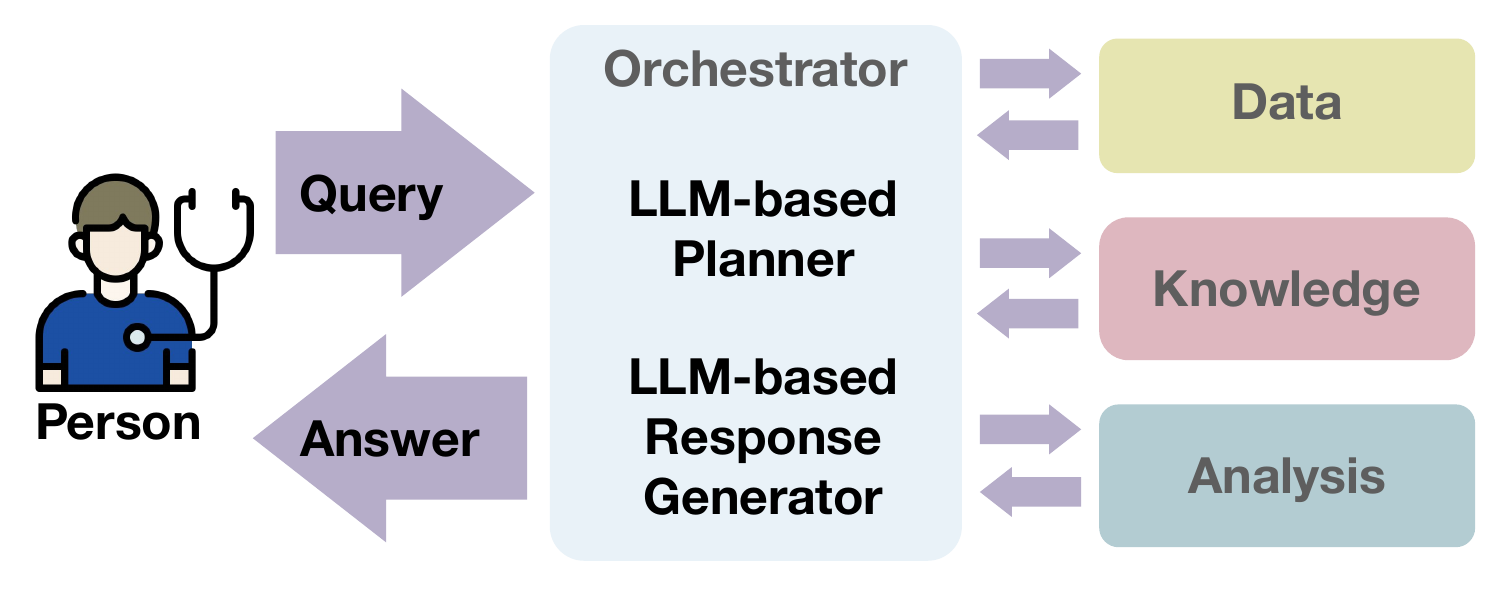}
    \caption{An overview of openCHA~\cite{abbasian2023conversational} framework.}
            \label{fig:introductionopenCHA}
\end{figure}

We believe that the integration of external knowledge bases with LLMs presents a robust solution to current challenges by enhancing the models' capacity to access reliable information. To achieve this, Conversational Health Agents (CHAs) can play an important role \cite{li2024personal}. CHAs are conversational systems that provide healthcare services, such as assistance and diagnosis, leveraging agents as the decision-making core. openCHA~\cite{abbasian2023conversational}, an open-source CHA framework, provides a flexible framework for implementing LLM-based diabetes management through its adept integration of diverse external sources. openCHA possesses the capability to integrate and infuse knowledge into chatbots, while also integrating external analytical methods for data analysis. Figure~\ref{fig:introductionopenCHA} shows an overview of this integration process, wherein an orchestrator -- powered by an LLM-based planner and an LLM-based response generator -- 
 engages with various sources to gather the required information. This collected information is then used to craft a response to the user's query. 

This paper presents a knowledge-based diabetes management system that is implemented through the utilization of LLM-powered CHAs. We customize the open-source openCHA framework for diet assessment by integrating external knowledge -- including the diet guidelines from the American Diabetes Association reports and the nutritional data from the Nutritionix knowledge base -- into LLMs. We also incorporate an analysis tool to precisely calculate the total daily nutritional intake and compare it against the guidelines. We evaluate the proposed approach with real-world diabetes-related questions focusing on daily meals in comparison to GPT4. The evaluation is centered on the assessment of the potential risk associated with the recommended dietary options.

\section{Method}

We develop an LLM-based CHA, designed to assess the risk associated with daily food intake according to recommended nutritional thresholds. We leverage openCHA \cite{abbasian2023conversational}, as a foundational framework, for our development. The proposed CHA aims to interact with users with diabetes, get their daily food intake information in the form of a conversation, and use a food knowledge base as well as guidelines to answer users in a more reliable way.

\begin{figure}[!t]
\includegraphics[width=\linewidth, trim={6.5cm 1cm 25cm 7cm},clip]{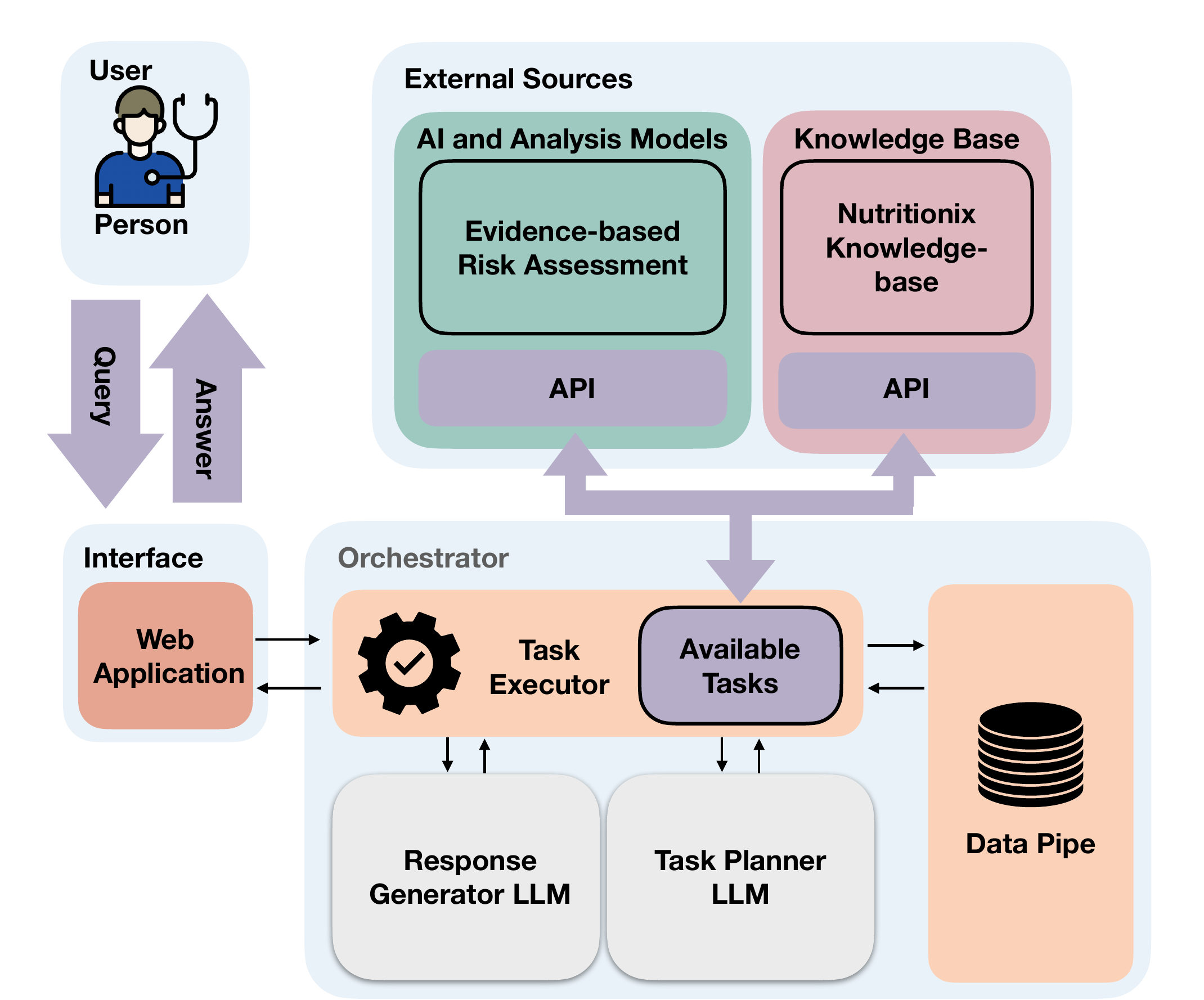}
\centering
\caption {LLM-based CHA for diabetes management enabled by the openCHA framework.} \label{fig:implementation}
\end{figure}

Our proposed CHA contains three main parts as Interface, Orchestrator, and External Sources (see Figure \ref{fig:implementation}). The Interface serves as a connection point between users and our framework, facilitating interactions through text using a web chat interface and relaying user queries to the Orchestrator.

The Orchestrator serves as the core component of the CHA and is responsible for problem-solving, planning, executing actions, and generating appropriate responses based on user queries. It operates on the principles of the Perceptual Cycle Model \cite{neisser1978perceiving} and involves perceiving, transforming, and analyzing the world (i.e., input query and metadata). It interacts with external sources, acquires information, performs data integration, and extracts insights. The Orchestrator comprises four major components: 1) Task Planner: to perform decision-making and planning, 2) Task Executor: to enable actuation and data conversion, 3) Data Pipe: to act as a repository for acquired metadata and intermediate data, 4) and Response Generator: to refine information and delivers conclusive responses. 
For the Orchestrator setup, we use the OpenAI's \cite{chatgptwebsite} GPT-3.5-turbo model as our base LLM, and the Tree of Thought \cite{yao2023tree} prompting technique for Planning.

External Sources are crucial for acquiring valuable information from diverse origins. Through the openCHA framework~\cite{abbasian2023conversational}, we integrate two primary external sources to the CHA as: 1) Knowledge Base and 2) AI and Analysis Models (see Figure \ref{fig:implementation}).

\textbf{Knowledge Base:} serves to fetch current healthcare information from credible sources. We developed a task focused on retrieving nutritional information from Nutritionix for various foods, enhancing the up-to-dateness of the responses. This task takes a food query as input and uses the NutritioniX API~\cite{nutritionX23} to retrieve a list of foods along with their corresponding nutrient information, based on standard measurements. This is achieved by leveraging the API's Natural Language for Nutrients feature, which allows for parsing of natural language requests and returning detailed nutrition information for common foods. 

\textbf{AI and Analysis Models:} provide data analysis, enabling the extraction of insights and associations. This is particularly valuable in intricate healthcare contexts where LLMs may not efficiently handle intensive computations. We develop a task focused on daily food risk assessment into the openCHA framework \cite{abbasian2023conversational}. This task follows established guidelines for daily nutrient intake to perform food risk assessments. The task also leverages the knowledge extracted from the NutritioniX using Knowledge Base developed task. This algorithm evaluates potential risks associated with food intake by comparing extracted nutritional data to recommended thresholds for diabetics, accounting for factors like carbohydrate content, sugar levels, protein amount, and dietary fiber.

\begin{figure*}[!t]
\includegraphics[width=\linewidth, trim={0cm 6cm 0cm 0cm},clip]{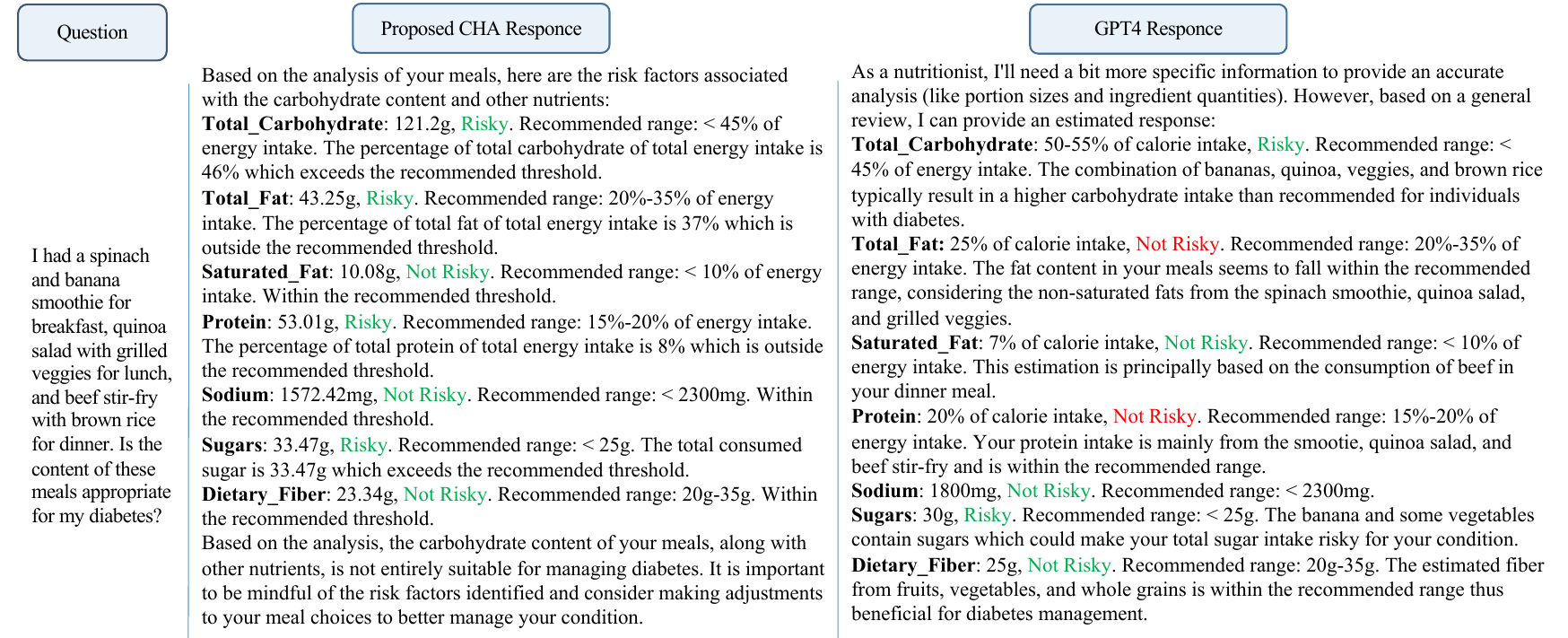}
\centering
\caption {A sample question and responses from the proposed CHA and GPT4. The green text indicates matching the identified risk with the ground truth, the red text indicates mismatching.} \label{fig:sample}
\end{figure*}

The threshold levels for the risk assessment algorithm are selected based on commonly recognized and accepted standards by medical associations for nutrient intake. According to an American Diabetes Association report \cite{20195.Diabetes-2019}, individuals suffering from Type 2 Diabetes should eat 20 to 35 grams (g) of fiber from raw vegetables and unprocessed grains per day and limit sodium intake to 2,300 milligrams (mg) per day, both recommendations aligning with those for the general population. Although there are no overarching recommended ranges for carbohydrate intake, based on \cite{Gray2019NutritionalDiabetes}, patients with Type 1 and type 2 diabetes in the US consumed about 45\% of their total energy intake through carbohydrates. While there is no concrete evidence about a certain range of protein intake aiding in glycemic control, we set protein intake to be 15-20\% of total calorie intake based on ADA standards of care \cite{20195.Diabetes-2019}. The recommended intake for fat is 20-35\% of total calories according to \cite{Millen20142013Nutritionists,Snell-Bergeon2009AdultsCalcium} while saturated fat is best restricted to less than 10\% of total calories \cite{VanHorn2008TheDisease}. Based on the American Heart Association (AHA) guidelines \cite{R.K.2009AMERICANASSOCIATION}, we set stringent restrictions on sugar intake due to its negative impact on diabetes with a maximum of 6 teaspoons or 25g of sugar.

\section{Results}

\begin{table}[!t]
    \centering
\caption{Sample evaluation for one question.}
\label{tab:sample}
\resizebox{\linewidth}{!}{
    \huge
    \begin{tabular}{|c|c|c|c|c|c|c|c|} \hline 
         &  Carbohydrate&  Fat&  Saturated Fat&  Protein&  Sodium&  Sugars& Dietary Fiber\\ \hline 
         Grount Truth&  R & R & NR & R & NR & R & NR \\ \hline 
         Proposed CHA&  R & R & NR & R & NR & R & NR \\ \hline
         GPT4&  R & NR & NR & NR & NR & R & NR \\ \hline 
        \multicolumn{8}{l}{R=Risky, NR=Not Risky}
    \end{tabular} 
}

\end{table}

In this section, we assess the effectiveness of our proposed CHA in evaluating food-related risks for individuals with diabetes. Our objective is to determine the CHA's ability to gauge the risk associated with daily food intake when the nutritional content deviates from recommended ranges. We compare the performance of our CHA, equipped with comprehensive food knowledge and analytical capabilities, with the OpenAI's GPT4 model \cite{chatgptwebsite}.

We collected 100 sample questions related to individuals inquiring about their daily meal intake. For the ground truth, we manually extract the nutritional data for each question and compute the overall nutritional values for parameters, such as fat, saturated fat, protein, carbohydrates, sodium, sugars, and dietary fiber. Subsequently, we evaluate the alignment of the nutrients with established guidelines. If any of these nutrients fall outside the recommended range, we categorize them as "Risky," while those within the range are labeled as "Not Risky." Ultimately, we generate a table of 100 rows and 7 columns (as our ground truth), representing Risky or Not Risky for each nutrient.

These questions are posed to both our CHA and GPT4 \cite{chatgptwebsite} in order to compare their responses with the ground truth. Figure \ref{fig:sample} provides an illustrative instance of a question and responses from the proposed CHA and GPT4.
The green text highlights the model's accurate identification of risk for specific nutrients based on the ground truth, while the red text indicates instances where the chatbot incorrectly assesses the risk for the corresponding nutrient. Table \ref{tab:sample} shows the ground truth values and summarizes the assessments of the two chatbots for each nutrient.

Similarly, we replicate this procedure for each set of 100 questions and then compare the quantity of correctly identified risks against the ground truth. For GPT4, we formulate a customized prompt to instruct GPT4 to estimate the nutritional information of the mentioned foods and assess their alignment with recommended guidelines and rules.

\begin{table}[!t]
    \centering
\caption{The risk assessment accuracy of the 100 sample questions.}
\label{tab:result}
\resizebox{\linewidth}{!}{
    \huge
    \begin{tabular}{|c|c|c|c|c|c|c|c|} \hline 
         &  Carbohydrate&  Fat&  Saturated Fat&  Protein&  Sodium&  Sugars& Dietary Fiber\\ \hline 
         GPT4&  49\% & 52\% & 68\% & 17\% & 73\% & 58\% & 46\% \\ \hline 
         Proposed CHA&  \textbf{84\%} & \textbf{94\%} & \textbf{99\%} & \textbf{90\%} & \textbf{95\%} & \textbf{91\%} & \textbf{92\%} \\ \hline
    \end{tabular} 
}
\end{table}

Table \ref{tab:result} indicates the responses to the 100 questions of both GPT4 and the proposed CHA. Our CHA outperforms GPT4 across all seven nutrients. This shows the significance of incorporating guidelines, knowledge bases (e.g., NutritioniX), and analytical tools into LLMs for health management tasks. It is worth noting that, in certain instances, the nutrient calculations for the proposed CHA bordered closely to the established guidelines, resulting in an increased error rate in some cases. For instance, when manually calculated, Carbohydrate content was found to be 44\%, which falls within the recommended range of less than 45\%, whereas our CHA calculated it to be 45\%. This might be due to minor variations in portion sizes or rounding of fractional numbers.

\section{Discussion}
The proposed CHA offers a significant degree of \textit{flexibility} for the integration of LLMs with external health data sources, knowledge bases, and analytical tools. This flexibility provides opportunities for mitigating hallucination problems while enhancing the personalization and reliability of the CHA. For instance, it allows for the incorporation of diverse elements such as food-related knowledge graphs, personal health biomarkers, individual demographics, and food preferences into the existing CHA framework, thereby enabling more refined and tailored recommendations.

Additionally, this development emphasizes a strong focus on \textit{explainability}. It enables users to inquire about the sequence of tasks and actions involved in generating a response. For example, when users seek information about specific food nutrients, data sources, or risk calculation methods, our CHA can provide insights into the underlying tasks and display results (in contrast to the state-of-the-art chatbots). This heightened level of transparency enhances the trustworthiness of the CHAs, fostering user confidence in the responses.

\section{Conclusion}

In this paper, we proposed an LLM-based CHA for diabetes management enabled by knowledge-infused LLMs. To achieve this, We leveraged openCHA for our development. We developed a nutrition information retrieval task to integrate the Nutritionix knowledge base into the CHA. Moreover, we developed a food risk assessment tool based on the American Diabetes Association dietary guidelines. We evaluated the effectiveness of our proposed CHA, in comparison to GPT4, in food-related risks for individuals with diabetes. Our findings showed that the proposed CHA outperforms the general-purpose GPT4 in responding to 100 diabetes-related queries regarding daily meal choices. This advancement highlights the potential of LLM-powered conversational agents in improving the accessibility and effectiveness of diabetes management, addressing a crucial aspect of healthcare for diabetic patients.

\bibliographystyle{ieeetr}
\bibliography{references}
\end{document}